\documentclass[letterpaper, 10 pt, conference]{ieeeconf}  

\IEEEoverridecommandlockouts                              

\usepackage{graphics} 
\usepackage{epsfig} 
\usepackage{booktabs}
\usepackage[table]{xcolor}

\usepackage{amsmath,amssymb,graphicx}
\usepackage{algorithm}
\usepackage{algpseudocode}
\usepackage{hyperref}

\title{\LARGE \bf
NavProbe: Evidence-Grounded Reasoning with Active Memory Retrieval for Zero-Shot Navigation}

\author{
Jingyang Liu$^{1}$,
Sujia Yao$^{1}$,
Jiayuan Gu$^{1}$,
and Lan Xu$^{1,\dagger}$%
\thanks{$^{1}$The authors are with the School of Information Science and
Technology, ShanghaiTech University, Shanghai 201210, China.
{\tt\small \{liujy2025, yaosj2024, gujy1, xulan1\}@shanghaitech.edu.cn}.}%
\thanks{$^{\dagger}$Corresponding author: Lan Xu.}
}

\begin{document}
\bstctlcite{IEEEexample:BSTcontrol}

\maketitle
\thispagestyle{empty}
\pagestyle{empty}


\begin{abstract}
Long-horizon navigation requires an agent to revise its intermediate objectives
as evidence accumulates.
Full visual histories are costly to process, while compact summaries may omit
details needed to reconsider earlier decisions.
We introduce NavProbe, a hierarchical zero-shot navigation agent that couples
a dynamic subgoal agenda with active evidence retrieval.
A compact index links summaries of visited places, transitions, and landmarks to their visual and geometric records.
When the current context is insufficient, a task executive retrieves targeted
evidence to generate, revise, or resolve subgoals.
Reusable conclusions are used to update the index, and a skill policy converts the revised task state into parameterized navigation actions.
NavProbe achieves 71.7\% SR and 55.8\% SPL on R2R-CE and 55.3\% SR and 38.6\% SPL on RxR-CE, outperforming strong zero-shot baselines.
It also achieves 79.3\% SR on HM3D-v2 ObjectNav, with qualitative real-robot demonstrations illustrating physical deployment. Code is available at
\url{https://github.com/liujy25/NavProbe}.
\end{abstract}


\section{Introduction}
\label{sec:introduction}

Pretrained vision-language models enable zero-shot navigation from route
instructions or object goals~\cite{navgpt,discussnav,hsgm,instructnav}.
Beyond selecting executable movements~\cite{vlnce}, an agent must continually
assess which intermediate objectives remain relevant and what has already
been accomplished.
In vision-and-language navigation (VLN), a turn made before the intended
landmark may be incorrectly marked complete.
In object-goal navigation (ObjectNav), a misidentified room may lead the agent
to pursue an unproductive search.
Correcting either decision requires a revisable task state and evidence that
may no longer be visible.

Access to navigation history presents a fundamental challenge for VLM reasoning. Retaining the full visual history preserves potentially useful evidence, but causes the context to accumulate redundant, stale, and currently irrelevant observations. Such extraneous information can interfere with multimodal reasoning,
as VLMs remain susceptible to contextual distractions and may incorporate
irrelevant visual cues as evidence~\cite{liu2026distractions}.
Compact memories reduce this burden, but may discard visual details
needed by later decisions or preserve earlier interpretations in place
of the underlying evidence~\cite{ren2026memlens}. Retrieval provides a middle ground by selectively exposing historical observations when they become relevant, as explored by ReMEmbR~\cite{remembr} and AgenticNav~\cite{agenticnav}. We focus on using such retrieved evidence to reconsider which objectives should be pursued, revised, or resolved during navigation.

We introduce NavProbe, a hierarchical VLM agent that couples a dynamic subgoal agenda with graph-structured episodic memory. Instruction clauses initialize the agenda in VLN, while intermediate search objectives are constructed online in ObjectNav. The episodic memory stores visual and geometric records of visited places, transitions, and landmarks. Compact entity summaries index these records, providing routine context and access to original evidence when needed.

NavProbe follows a \emph{formulate--probe--verify--act} cycle.
A task executive assesses the dynamic agenda using the current observation, task state,
and memory index.
Unresolved questions trigger targeted retrieval, whose conclusions can generate
new subgoals, revise priorities, or reopen previously resolved objectives.
Reusable conclusions are fed back to update entity knowledge.
A skill policy combines the revised task state and retrieval conclusions with
the current observation to select a parameterized navigation action.
Execution extends episodic memory for subsequent decisions.

Our contributions are:
\begin{itemize}
    \item We introduce evidence-grounded task-state reasoning that uses active retrieval to generate, revise, and resolve navigation subgoals.
    
    \item We develop graph-structured episodic memory with a compact index of entity knowledge linked to retrievable visual and geometric records, together with consolidation of reusable retrieval-derived conclusions for subsequent decisions.
    
    \item We evaluate NavProbe on R2R-CE and RxR-CE, where it achieves the highest SR among the compared zero-shot methods. Ablations and retrieval analysis examine the contributions of task-state memory, active retrieval, and knowledge consolidation, while ObjectNav and real-robot experiments assess the framework beyond the main VLN setting.
\end{itemize}

\section{Related Work}
\label{sec:relatedwork}
\subsection{Zero-Shot VLN and Navigation Recovery}

Recent training-free VLN methods use pretrained vision-language models
for high-level reasoning and navigation decision-making.
NavGPT~\cite{navgpt} represents observations and navigation history in language,
while DiscussNav~\cite{discussnav} coordinates multiple model-based experts
for instruction understanding, perception, and decision-making.
InstructNav~\cite{instructnav} grounds language-conditioned
reasoning through multi-source value maps.
Other methods explicitly maintain the state of instruction execution.
Open-Nav~\cite{opennav} incorporates progress estimation into its reasoning
process, CA-Nav~\cite{canav} models navigation as sequential satisfaction of
sub-instruction constraints, and GC-VLN~\cite{gcvln} represents instructions
as graph-structured spatial constraints.

Navigation recovery has also been studied as a way to improve robustness
to errors during long-horizon execution.
Earlier VLN methods use progress estimates or trajectory-level search to
determine when and where to backtrack~\cite{regretful_agent}.
More recent training-free systems incorporate recovery directly into
VLM-based navigation.
SmartWay~\cite{smartway} combines history-aware reasoning with backtracking
to the last state, while GC-VLN~\cite{gcvln} performs backtracking over its
constraint-based navigation tree.
Uni-LaViRA~\cite{unilavira} introduces Second Chance Backtrack, which selects
a previously visited waypoint and uses the failed sub-trajectory to support
the subsequent decision.
These approaches provide different mechanisms for recovering from navigation
errors.
NavProbe instead focuses on how the historical evidence supporting such
decisions is represented and selectively accessed during navigation.

\subsection{Navigation Memory and Evidence Retrieval}

Long-horizon navigation requires memory representations that preserve
task-relevant observations and spatial information over extended execution.
SpaceVLN~\cite{spacevln} maintains waypoint and landmark memories for
progress localization and spatial reasoning.
ReMEmbR~\cite{remembr} associates visual observations with spatial and temporal
information, and RAVEN~\cite{raven} stores visual representations together
with pose and time for semantic, spatial, and temporal retrieval.

Memory representation alone does not determine what information enters the
reasoning context.
AgenticNav~\cite{agenticnav} combines a compact trajectory map with an
on-demand recall tool for revisiting past visual observations.
HAM-VLN~\cite{hamvln} retains recent waypoints in context and retrieves older
history using relevance, recency, salience, and topological expansion.
STaR~\cite{star} uses task-conditioned aggregation to select multimodal memories,
while CMMR-VLN~\cite{cmmrvln} retrieves stored multimodal navigation evidence
for long-horizon reasoning.
NavProbe maintains compact, updatable entity knowledge as an index over
multimodal records associated with visited places, executed transitions, and
task-relevant landmarks.
The agent can query specific entities and evidence fields when the underlying
records are needed, while retrieval-derived conclusions can be consolidated
into entity knowledge for subsequent decisions.

\section{Method}
\label{sec:method}

\begin{figure*}[t]
    \centering
    \vspace*{3mm}
    \includegraphics[width=0.85\textwidth]{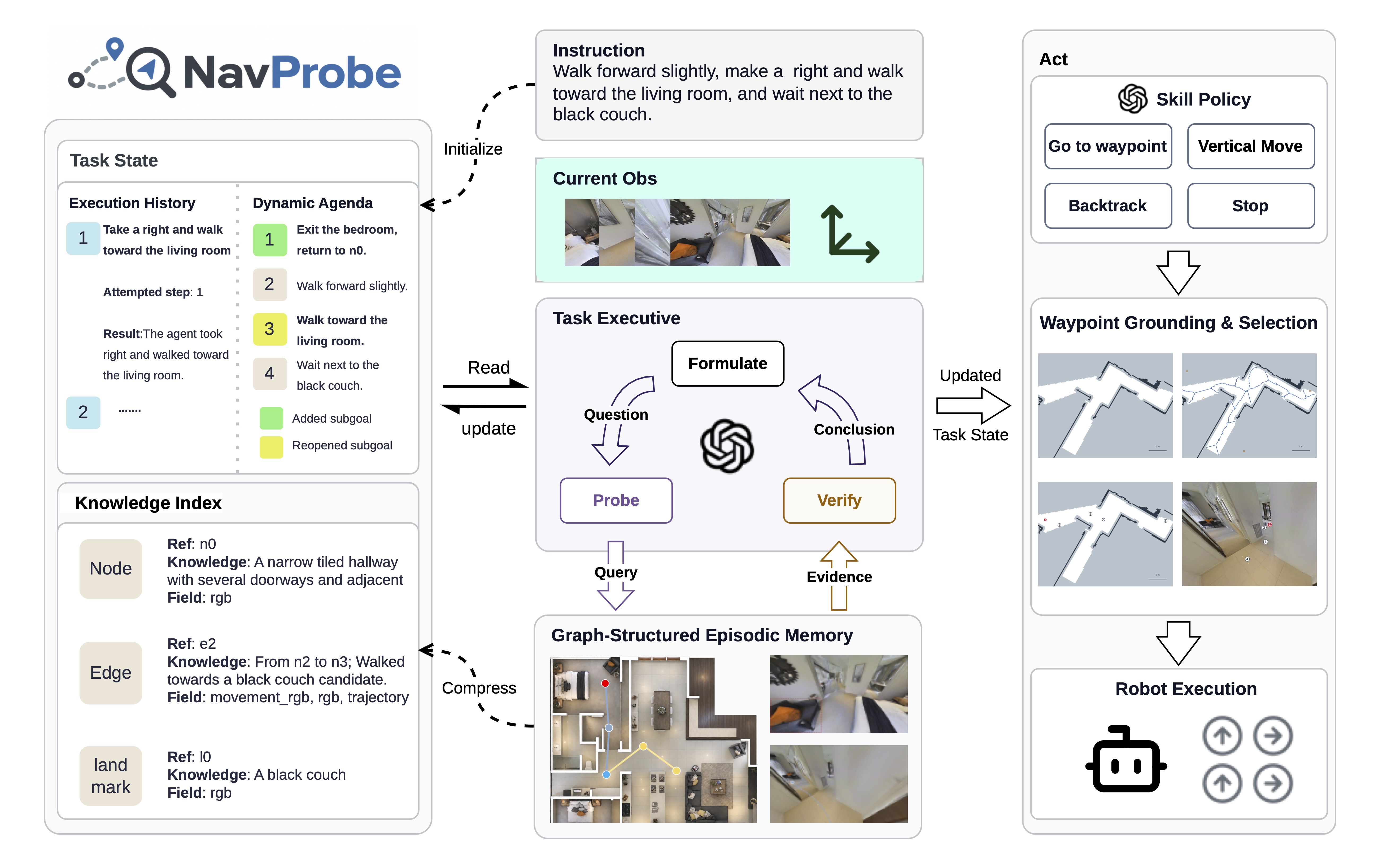}
    \caption{
    NavProbe's formulate--probe--verify--act cycle.
    The task executive maintains a dynamic subgoal agenda using the current
    observation, task-state memory, and compact episodic-memory index.
    It retrieves historical evidence when needed to generate, revise, or resolve
    subgoals, and consolidates reusable conclusions into entity knowledge.
    The skill policy combines the updated task state and retrieval conclusions
    with the current observation to select a parameterized navigation action. 
    Geometric routines execute the action, producing observations and trajectories
    for subsequent decisions.
    }
    \label{fig:pipeline}
\end{figure*}

\subsection{NavProbe Overview}
\label{sec:overview}

Given a task goal $g$, expressed as a navigation instruction or an object
target, NavProbe maintains intermediate objectives and selects actions until
termination.
A hierarchical VLM agent operates through two role-specialized modules:
a \emph{task executive} for subgoal management and evidence retrieval, and a
\emph{skill policy} for action selection.
The modules use separate prompts and contexts.
Memory queries and task-state updates operate within a decision step;
navigation actions connect the agent to the environment.

The decision cycle follows \emph{formulate--probe--verify--act}.
At step $t$, the executive reads the current observation $O_t$, dynamic task
state $\mathcal{T}_t$, and compressed memory index $\mathcal{I}_t$.
It formulates relevant subgoals, assesses progress toward existing ones, and
identifies questions that need further evidence.
When needed, it queries historical records, interprets them and uses the conclusion to revise the agenda, repeating this reasoning loop without moving the robot.
When no further retrieval is requested or the budget is exhausted, it passes
the updated task state and evidence conclusions to the skill policy.
The policy selects a navigation action using this output and the current
observation.
Execution extends episodic memory with new observations and trajectories,
while consolidation preserves reusable evidence conclusions in the index.

\subsection{Hierarchical Navigation Memory}
\label{sec:memory}

NavProbe distinguishes three roles for memory: the dynamic state of the task,
persistent episodic records of navigation, and compact entity knowledge that
indexes those records.
The executive routinely reads the task state and index; historical images
enter its context only when explicitly retrieved.

\subsubsection{Task State and Dynamic Subgoal Management}
\label{sec:progress_memory}

Task-state memory represents the agent's current intermediate objectives and
the outcomes of earlier decisions:
\begin{equation}
    \mathcal{T}_t=(\mathcal{A}_t,\mathcal{H}_t),
    \qquad \mathcal{A}_t=[g_{1,t},\ldots,g_{K_t,t}],
\end{equation}
where $\mathcal{A}_t$ is the subgoal agenda and $\mathcal{H}_t$ is execution
history.
Both the content and size $K_t$ of the agenda can change during navigation.
Each entry $g_{i,t}$ contains a natural-language objective, its execution
status, and a set of verification predicates $C_{i,t}$.
The executive can introduce new subgoals, refine their descriptions,
reprioritize their pursuit, or remove objectives that are satisfied or no
longer useful.
Agenda updates remain grounded in the original task goal and any ordering
constraints specified by the instruction.

For VLN, explicit instruction clauses provide initial subgoals that can be
refined as the environment is observed.
For ObjectNav, the target specifies the task goal, while the executive
constructs intermediate search objectives online, such as exploring a likely
room, inspecting a candidate region, or revisiting an earlier observation.
Thus, a search objective can be completed without the overall task being
complete: inspecting a countertop may establish that the target is absent
and motivate searching elsewhere.

A verification predicate pairs an evidence-checkable proposition with a
\texttt{confirmed} or \texttt{unconfirmed} status.
Predicates capture observations and relations relevant to pursuing or resolving
a subgoal, such as whether the referenced landmark has been identified.
These predicate sets are initially empty and can be extended or revised as evidence
accumulates.
Subgoal resolution is assessed against the full objective and available evidence.
History records previously pursued subgoals and their outcomes, including
completion or abandonment, with execution summaries and spatial references
such as the start and end topological nodes.
It supports reconsideration of earlier decisions: contradictory evidence can
return a previously resolved objective to the agenda.

\subsubsection{Graph-Structured Episodic Records}
\label{sec:episodic_memory}

A topological graph $\mathcal{G}_t=(V_t^P,E_t)$ organizes visited places and
executed transitions, with task-relevant landmark entities $V_t^L$ associated
with the navigation history.
Place nodes store panoramic observations; transition edges record executed
trajectories; and landmark records contain detections and estimated locations.
These visual and geometric records preserve the evidence behind earlier
observations and actions.
A separate occupancy map $M_t$ records free, occupied, and unexplored regions
for spatial reasoning and waypoint generation.

\subsubsection{Entity Knowledge and the Compressed Index}

Each place, transition, or landmark entity $e$ has compact textual knowledge
$\mathcal{K}_e^E$ linked to its episodic records.
This semantic layer is initialized from place observations, the reasoning
accompanying executed transitions, or landmark detections, respectively.
It is subsequently revised through evidence consolidation.
The summaries and available evidence fields form the index
\begin{equation}
    \mathcal{I}_t =
    \left\{(e,\mathcal{K}_e^E,\mathcal{A}_e)
    \mid e\in V_t^P\cup E_t\cup V_t^L\right\},
\end{equation}
where $\mathcal{A}_e$ identifies the retrievable fields and their record
references.
The index provides the executive with a compact interpretation of navigation
history and pointers to the evidence needed to check that interpretation.
Task state and entity knowledge serve complementary purposes:
$\mathcal{T}_t$ records the current agenda and its execution history, while
entity summaries retain reusable knowledge about places, transitions, and
landmarks.

\subsection{Evidence-Grounded Task-State Updates}
\label{sec:planning_agent}

\subsubsection{Formulating Subgoals and Probing Memory}
\label{sec:active_retrieval}

The task executive begins with $(O_t,\mathcal{T}_t,\mathcal{I}_t)$ and
considers which objectives to pursue, revise, or resolve.
If a task-state decision requires further evidence, it formulates a query
$q_t^r$ in reasoning round $r$.
Queries may concern whether a subgoal has been satisfied, whether a proposed
search region remains relevant, or whether an earlier observation supports
introducing a new objective.
For example, the executive may check whether a turn occurred after the required
landmark or whether a previously visited room warrants another inspection.
The executive uses the index to select the relevant entities and evidence
fields, then invokes the memory retrieval tool:
\begin{equation}
\begin{aligned}
    R_t^r &= \{(e_j,\mathcal{F}_j)\}_{j=1}^{m_r},\\
    D_t^r &= \operatorname{Retrieve}(R_t^r),
\end{aligned}
\end{equation}
where \(m_r\) is the number of entity-field pairs selected in reasoning round \(r\), $e_j\in V_t^P\cup E_t\cup V_t^L$ and
$\mathcal{F}_j\subseteq\mathcal{A}_{e_j}$.
The query determines what to inspect and how to interpret it; the selected
entity--field references specify the records returned by the tool.
The batch $D_t^r$ contains the requested historical images and geometric records.
Retrieval is triggered by an unresolved task-state question.
When the initial context is sufficient, the executive updates the agenda and
proceeds to action selection without retrieving historical records.

\subsubsection{Evidence Verification and Task-State Revision}

The executive inspects $D_t^r$ and produces a textual conclusion $c_t^r$
that states whether and how the evidence supports or contradicts $q_t^r$.
The conclusion informs the generation, revision, or resolution of subgoals,
updates to their verification predicates and execution history, and any further
evidence request.
For VLN, evidence that a turn preceded the required landmark can reopen the
corresponding objective.
For ObjectNav, a retrieved panorama showing kitchen fixtures near an overlooked
doorway may support a new subgoal to revisit and inspect that region.
Retrieval can therefore change which objectives the agent pursues as well as
its assessment of progress toward them.

Within each decision step, a temporary workspace retains the queries,
selected record references, and evidence conclusions.
Raw retrieved records remain in the active context only during inspection;
subsequent rounds use their conclusions and references, while the originals
remain available in episodic memory.
A cumulative bird's-eye-view (BEV) representation preserves the spatial context
of the inspected entities and trajectories for reasoning and action selection.

The loop ends when the executive requests no further evidence or reaches the
per-step budget $R_{\max}$.
At the budget limit, the final batch is interpreted before the task-state
updates are finalized, even if some questions remain unresolved.
The executive's output consists of the updated state $\widehat{\mathcal{T}}_t$,
a task-state assessment, retrieval conclusions, and cumulative BEV when available.
The assessment explains which objectives remain relevant and how the evidence
informs their pursuit.
The revised task state is retained for subsequent navigation decisions.

\subsubsection{Memory Consolidation}
\label{sec:knowledge_consolidation}

After retrieval, reusable evidence conclusions are consolidated into the
corresponding entity summaries.
This operation, implemented by the Entity Knowledge Manager (EKM), takes the
textual query--reference--conclusion records and existing entity knowledge as
input.
The record references associate each conclusion with its source evidence and
the entities to update.
Consolidation adds new information, revises existing entries, or removes claims
contradicted by the inspected evidence, without reopening the retrieved images.
The updated summaries enter $\mathcal{I}_t$ at subsequent decisions, allowing
the executive to reuse established conclusions when managing the agenda.

\subsection{Act: Skill Selection and Geometric Execution}
\label{sec:skill_execution}

The skill policy maps the current observation and the executive's output
$Y_t$ to a parameterized action:
\begin{equation}
    a_t=(d_t,\theta_t)=\pi_{\mathrm{skill}}(O_t,Y_t).
\end{equation}
Here, $Y_t$ contains $\widehat{\mathcal{T}}_t$, the task-state assessment,
evidence conclusions, and available spatial context described above.
The updated agenda specifies the objectives currently under consideration,
while execution history informs the choice of what to pursue next.
The action type $d_t$ is \texttt{GoToWaypoint}, \texttt{VerticalMove},
\texttt{Backtrack}, or the terminal action \texttt{Stop}; $\theta_t$ contains
its arguments, such as a waypoint label or a visited node.
The policy receives textual conclusions from retrieved evidence
together with the current observation.
Geometric routines translate the selected action into low-level motion.

\subsubsection{Waypoint Grounding}

For \texttt{GoToWaypoint} and \texttt{VerticalMove}, we follow HSGM's visual
waypoint formulation~\cite{hsgm} and extend waypoint generation with
Frontier--Skeleton Sampling (FSS).
Given $M_t$, FSS samples frontier anchors along explored--unexplored boundaries
and skeleton anchors on the medial axis of navigable free space.
A collision-free path is computed toward each anchor, and a visible waypoint
is selected along it subject to distance and occlusion constraints.
Candidates receive consistent labels in the current RGB observation and BEV.
The skill policy selects a label using the executive's task state and evidence
conclusions, and the corresponding geometric waypoint is passed to the
low-level navigation controller.
For \texttt{VerticalMove}, the detected stair region is temporarily treated
as traversable before applying FSS.

\subsubsection{Recovery, Termination, and Memory Updates}

For \texttt{Backtrack}, the policy selects a visited place node for route
recovery and subsequent replanning.
For \texttt{Stop}, the robot retains its pose or performs final local
positioning toward the expected destination before terminating.
Executed movements contribute new observations and trajectories to episodic
memory.
These records, their entity summaries, and the retained task state provide
the context for the next formulate--probe--verify--act cycle.

\begin{table*}[t]
    \centering
    \vspace*{3mm}
    \caption{
    Main results on R2R-CE and RxR-CE val-unseen.
    Bold and underlined values denote the best and second-best results,
    respectively, within the zero-shot block.
    $^\dagger$ denotes our rerun of Uni-LaViRA using GPT-5.5
    on the R2R-CE subset.
    }
    \label{tab:main_results}
    \small
    \setlength{\tabcolsep}{6pt}
    \begin{tabular}{l|cccc|cccc}
        \toprule
        \textbf{Method}
        & \multicolumn{4}{c|}{\textbf{VLN-CE R2R Val-Unseen}}
        & \multicolumn{4}{c}{\textbf{VLN-CE RxR Val-Unseen}} \\
        & \textbf{NE}$\downarrow$ & \textbf{OSR}$\uparrow$ & \textbf{SR}$\uparrow$ & \textbf{SPL}$\uparrow$
        & \textbf{NE}$\downarrow$ & \textbf{SR}$\uparrow$ & \textbf{SPL}$\uparrow$ & \textbf{nDTW}$\uparrow$ \\
        \midrule
        \rowcolor[gray]{0.85}
        \multicolumn{9}{c}{\textbf{Supervised Learning (Training Method)}} \\
        StreamVLN~\cite{streamvln}    & 4.98 & 64.2 & 56.9 & 51.9 & 6.22 & 52.9 & 46.0 & 61.9 \\
        NavFoM~\cite{navfom}       & 5.01 & 64.9 & 56.2 & 51.2 & 5.51 & 57.4 & 49.4 & 60.2 \\
        OmniNav~\cite{omninav}      & 3.74 & 74.6 & 69.5 & 66.1 & 3.77 & 73.6 & 62.0 & -- \\
        InternVLA-N1~\cite{internvlan1} & 4.83 & 63.3 & 58.2 & 54.0 & 5.91 & 53.5 & 46.1 & 65.3 \\
        ABot-N0~\cite{abotn0}      & 3.78 & 70.8 & 66.4 & 63.9 & 3.83 & 69.3 & 60.0 & -- \\
        \midrule
        \rowcolor[rgb]{0.85,0.95,0.99}
        \multicolumn{9}{c}{\textbf{Zero-Shot (Training-Free)}} \\
        InstructNav~\cite{instructnav}  & 6.89 & 47.0 & 31.0 & 24.0 & -- & -- & -- & -- \\
        Open-Nav~\cite{opennav}     & 6.70 & 23.0 & 19.0 & 16.1 & -- & -- & -- & -- \\
        CA-Nav~\cite{canav}       & 7.58 & 48.0 & 25.3 & 10.8 & 10.4 & 19.0 & 6.0 & 13.5 \\
        SmartWay~\cite{smartway}     & 7.01 & 51.0 & 29.0 & 22.5 & -- & -- & -- & -- \\
        GC-VLN~\cite{gcvln}       & 7.30 & 41.8 & 33.6 & 16.3 & 8.80 & 33.8 & 13.8 & -- \\
        HiMemVLN~\cite{himemvln}     & 6.65 & 36.0 & 30.0 & 26.9 & -- & -- & -- & -- \\
        Three-Step Nav~\cite{threestep} & 5.87 & 39.0 & 34.0 & 29.1 & 9.21 & 22.0 & 16.1 & 45.7 \\
        Uni-LaViRA~\cite{unilavira}
        & \underline{3.66} & \underline{73.7} & 60.7 & \underline{47.7}
        & \underline{6.48} & \underline{51.3} & \underline{34.0} & \underline{53.7} \\
        Uni-LaViRA\,(GPT-5.5)$^\dagger$
        & 4.02 & 73.3 & \underline{61.7} & 45.0
        & -- & -- & -- & -- \\
        \textbf{NavProbe (Ours)}
        & \textbf{3.45} & \textbf{78.0} & \textbf{71.7} & \textbf{55.8}
        & \textbf{5.39} & \textbf{55.3} & \textbf{38.6} & \textbf{53.9} \\
        \bottomrule
    \end{tabular}
\end{table*}

\section{Experiments}
\label{sec:experiments}
\subsection{Experimental Setup}
\label{sec:experimental_protocol}

\noindent\textbf{Environments and Datasets.}
We evaluate on R2R-CE~\cite{vlnce} and
RxR-CE~\cite{vlnce,rxr} for vision-and-language navigation,
and HM3D-v2~\cite{hm3d} for object-goal navigation.
The two VLN-CE benchmarks use Matterport3D environments in
Habitat~\cite{matterport3d,habitat}.
R2R-CE contains relatively concise English route instructions, whereas
RxR-CE contains longer instructions with denser grounding to reference
trajectories.
Following Uni-LaViRA~\cite{unilavira}, we evaluate NavProbe on the same
stratified 100-episode subset for each benchmark. Results for
prior methods are their originally reported values unless otherwise noted.
Each NavProbe configuration is evaluated over three independent runs,
and we report the mean across runs.
These results refer to the evaluated subsets rather than the full
benchmark splits.

\noindent\textbf{Evaluation Metrics.}
For R2R-CE, we report Navigation Error (NE), Oracle Success Rate (OSR),
Success Rate (SR), and Success weighted by Path Length
(SPL)~\cite{spl}.
NE is the final geodesic distance to the goal; OSR records whether the
trajectory enters the goal region at any point; SR requires successful
termination; and SPL additionally accounts for path efficiency.
For RxR-CE, we report NE, SR, SPL, and normalized Dynamic Time Warping
(nDTW)~\cite{ndtw}, which measures alignment with the reference trajectory.
For HM3D-v2 ObjectNav, we report SR and SPL.
The configured success radii are $3.0\,\mathrm{m}$ for VLN-CE and
$1.0\,\mathrm{m}$ for HM3D-v2 ObjectNav.
Lower NE and higher values of the other metrics indicate better performance.

\noindent\textbf{Implementation Details.}
All VLM reasoning calls use GPT-5.5 with role-specific prompts.
Landmarks are detected using YOLO-World~\cite{yoloworld} with a
confidence threshold of 0.6.
Simulator RGB-D observations have a resolution of $960\times540$
and a $135^{\circ}$ horizontal field of view.
The camera is mounted at a height of $1.2\,\mathrm{m}$ with a
$-30^{\circ}$ pitch, and images passed to the VLM are resized to fit
within $640\times360$.
At each navigation decision, the agent constructs a four-view panorama
at relative headings of $0^{\circ}$, $90^{\circ}$, $180^{\circ}$,
and $270^{\circ}$.
Metric waypoints selected through the navigation skills are executed
using Habitat-Lab's \texttt{ShortestPathFollower}~\cite{habitat}.
Waypoint selection remains the
responsibility of NavProbe.
The per-decision retrieval budget is $R_{\max}=6$.
Each episode is limited to 40 navigation decisions and 500 simulator
actions, with a forward step size of $0.25\,\mathrm{m}$.

\subsection{Main Results}
\label{sec:main_results}

\noindent\textbf{VLN-CE Results.}
Table~\ref{tab:main_results} compares NavProbe with recent navigation methods
on R2R-CE and RxR-CE. We include representative trained methods for
performance context and a broader set of zero-shot methods for direct comparison.
On R2R-CE, NavProbe achieves 71.7\% SR and 55.8\% SPL, with
an NE of $3.45\,\mathrm{m}$ and an OSR of 78.0\%.
Our Uni-LaViRA rerun with GPT-5.5 as the high-level reasoning model
achieves 61.7\% SR and 45.0\% SPL.
NavProbe therefore exceeds this rerun by 10.0 percentage points in
SR and 10.8 points in SPL.
This comparison matches the high-level reasoning model, but does not
isolate individual architectural components; their contributions are
examined through the ablations below.

On RxR-CE, NavProbe achieves 55.3\% SR and 38.6\% SPL, compared
with Uni-LaViRA's reported 51.3\% SR and 34.0\% SPL.
NE decreases from $6.48\,\mathrm{m}$ to $5.39\,\mathrm{m}$,
while nDTW is similar at 53.9 versus 53.7.
The observed improvements are therefore primarily in successful
termination, SPL, and final goal proximity, rather than a substantial
change in trajectory alignment.
Together, these results establish the end-to-end performance of the
complete system on the evaluated VLN-CE subsets.

\noindent\textbf{ObjectNav Results.}
Table~\ref{tab:hm3dv2_results} extends the evaluation to object-goal
navigation on HM3D-v2.
NavProbe achieves 79.3\% SR, the highest listed value and 1.6 percentage
points above Uni-LaViRA.
Its SPL is 43.8\%, compared with Uni-LaViRA's 46.1\%.
Thus, NavProbe has the higher SR, while Uni-LaViRA retains
the higher SPL.

\begin{table}[t]
    \centering
    \caption{
    Main results on HM3D-v2 ObjectNav.
    Bold and underlined values denote the best and second-best results,
    respectively.
    }
    \label{tab:hm3dv2_results}
    \small
    \setlength{\tabcolsep}{10pt}
    \begin{tabular}{l|cc}
        \toprule
        \textbf{Method} & \textbf{SR}$\uparrow$ & \textbf{SPL}$\uparrow$ \\
        \midrule
        VLFM~\cite{vlfm}        & 62.6 & 31.0 \\
        InstructNav~\cite{instructnav} & 58.0 & 20.9 \\
        ApexNav~\cite{apexnav}     & 76.2 & 38.0 \\
        DSCD-Nav~\cite{dscdnav}    & 73.0 & 38.7 \\
        ReMemNav~\cite{rememnav}    & 67.8 & 36.6 \\
        Uni-LaViRA~\cite{unilavira}  & \underline{77.7} & \textbf{46.1} \\
        \textbf{NavProbe (Ours)} & \textbf{79.3} & \underline{43.8} \\
        \bottomrule
    \end{tabular}
\end{table}

\subsection{Ablation Studies}
\label{sec:ablation}

Table~\ref{tab:ablation} examines three complementary roles in NavProbe:
maintaining an explicit execution state, accessing historical evidence,
and retaining reusable conclusions for later decisions.
The Full Graph variant additionally evaluates the choice of historical
context supplied for reasoning.

\begin{figure*}[t]
    \centering
    \vspace*{3mm}
    \includegraphics[width=0.85\textwidth]{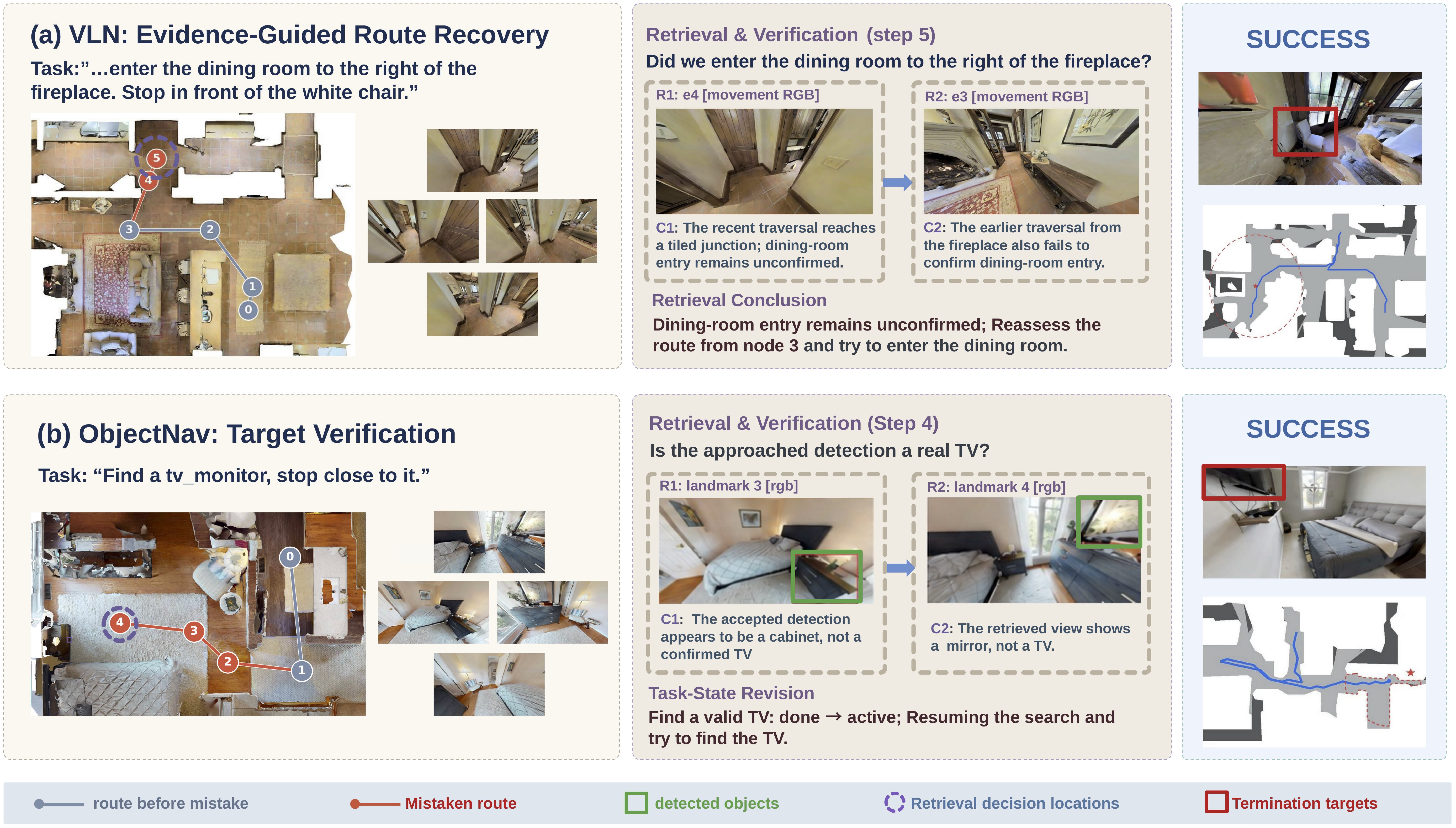}
     \caption{
      Qualitative results of retrieval-supported recovery.
      (a) In VLN, retrieved transition evidence guides route reassessment,
      followed by successful navigation to the dining room.
      (b) In ObjectNav, retrieved landmark evidence prompts renewed
      search after a false TV detection, leading to successful termination
      near a real TV.
      }
    \label{fig:result_vis}
\end{figure*}

\noindent\textbf{Task-State Memory.}
In \emph{w/o Task-State Memory}, the agent reasons from the
instruction at each decision without carrying structured task-state
memory across steps.
SR decreases from 71.7\% to 63.3\%, SPL decreases from 55.8\%
to 51.4\%, and NE increases from $3.45\,\mathrm{m}$ to
$3.94\,\mathrm{m}$.
The 8.4-point SR decrease supports maintaining an explicit execution
state across decisions, rather than reconstructing progress from the
instruction at every step.
This ablation evaluates task-state memory as a whole, including
subgoal statuses, verification predicates, and execution history.

\noindent\textbf{Memory Consolidation (EKM).}
In \emph{w/o EKM}, evidence retrieval remains enabled, but retrieval
conclusions are not consolidated into entity knowledge for subsequent
decisions.
Relative to full NavProbe, SR decreases from 71.7\% to 65.0\%
and SPL from 55.8\% to 51.1\%, while NE increases from
$3.45\,\mathrm{m}$ to $4.20\,\mathrm{m}$.
This comparison supports the contribution of updating persistent
entity summaries with retrieval-derived conclusions, beyond using
those conclusions within the current decision.

\begin{table}[h]
    \centering
    \caption{Ablation results on R2R-CE.}
    \label{tab:ablation}
    \small
    \setlength{\tabcolsep}{3.4pt}
    \begin{tabular}{l|cccc}
        \toprule
        \textbf{Variant}
        & \textbf{NE} $\downarrow$
        & \textbf{OSR} $\uparrow$
        & \textbf{SR} $\uparrow$
        & \textbf{SPL} $\uparrow$ \\
        \midrule
        \textbf{NavProbe\,(Ours)}
        & \textbf{3.45}
        & \textbf{78.0}
        & \textbf{71.7}
        & \textbf{55.8} \\
        w/o Task-State Memory
        & 3.94 & 72.0 & 63.3 & 51.4 \\
        w/o EKM
        & 4.20 & 67.7 & 65.0 & 51.1 \\
        w/o Retrieval \& EKM
        & 4.65 & 69.3 & 57.3 & 43.1 \\
        Full Graph
        & 3.83 & 70.3 & 62.0 & 48.8 \\
        \bottomrule
    \end{tabular}
\end{table}

\noindent\textbf{Active Evidence Retrieval.}
Disabling retrieval also disables EKM, which operates on retrieval
conclusions; this variant is denoted \emph{w/o Retrieval \& EKM}.
To evaluate retrieval separately from consolidation, we compare this
variant with \emph{w/o EKM}.
With consolidation disabled in both variants, enabling retrieval raises
SR from 57.3\% to 65.0\% and SPL from 43.1\% to 51.1\%, while
reducing NE from $4.65\,\mathrm{m}$ to $4.20\,\mathrm{m}$.
These differences support access to the underlying visual and geometric
records beyond reasoning from compact entity knowledge alone.
The comparison between full NavProbe and \emph{w/o Retrieval \& EKM}
reflects the combined contribution of retrieval and consolidation,
not retrieval in isolation.

\noindent\textbf{Selective Historical Context.}
\emph{Full Graph} replaces selective entity--field retrieval with
the corresponding full-graph history in the reasoning context.
This variant achieves 62.0\% SR and 48.8\% SPL, compared with
71.7\% and 55.8\% for NavProbe, and has an NE of
$3.83\,\mathrm{m}$ rather than $3.45\,\mathrm{m}$.
The result favors query-directed evidence selection over the tested
full-graph context configuration.

\subsection{Retrieval Analysis}
\label{sec:retrieval_analysis}

We analyze how NavProbe uses episodic evidence during the 600
VLN-CE episode executions: 100 episodes per benchmark, repeated
three times on each of R2R-CE and RxR-CE.
Statistics include both successful and unsuccessful executions.
We pool decision- and query-level counts across the two benchmarks
before computing frequencies, rather than averaging per-episode rates.

\begin{figure*}[t]
    \centering
    \vspace*{1.5mm}
    \includegraphics[width=0.85\textwidth]{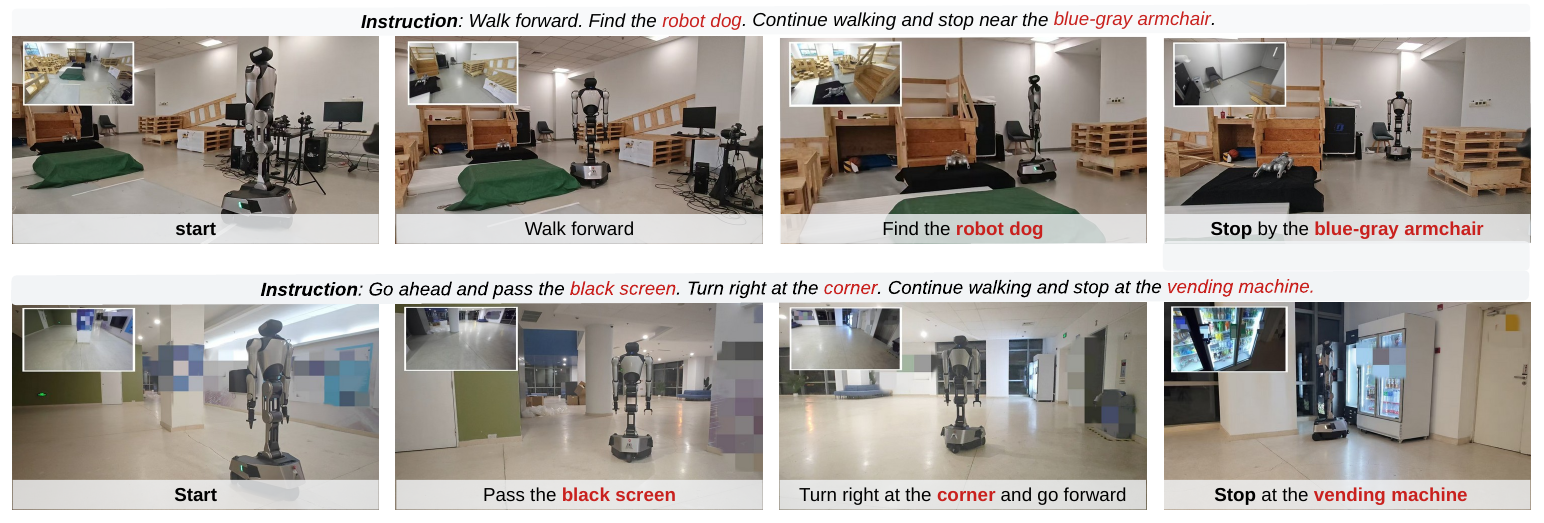}
    \caption{
    Real-world deployment of NavProbe on a Galaxea R1Pro.
    Each row shows key execution stages and corresponding onboard observations
    for a multi-stage navigation instruction.
    }
    \label{fig:real_world}
\end{figure*}

\noindent\textbf{Retrieval Usage.}
Of 4,554 navigation decisions, 2,127 invoke retrieval (46.71\%).
The first decision in each execution has no prior navigation history;
excluding these 600 decisions gives a retrieval rate of 53.79\%
over the remaining 3,954 decisions.
At decisions that invoke retrieval, the task executive performs an average of
1.82 retrieval rounds.
Historical evidence is therefore inspected at a subset of eligible
decisions rather than automatically at every step.

\begin{table}[h]
    \centering
    \caption{
    Semantic dimensions of 3,862 retrieval queries pooled across
    R2R-CE and RxR-CE.
    Labels are non-exclusive.
    }
    \label{tab:retrieval_query_dimensions}
    \small
    \setlength{\tabcolsep}{6pt}
    \begin{tabular}{@{}lrr@{}}
        \toprule
        \textbf{Query dimension} & \textbf{Queries} & \textbf{Share} \\
        \midrule
        Action execution
        & 2,643 & 68.44\% \\
        Spatial/route relation
        & 2,152 & 55.72\% \\
        Landmark/target identification
        & 1,122 & 29.05\% \\
        \midrule
        $\geq 2$ dimensions
        & 1,863 & 48.24\% \\
        \bottomrule
    \end{tabular}
\end{table}

\noindent\textbf{Questions Addressed by Retrieval.}
We annotate each retrieval query along three non-exclusive semantic
dimensions: \emph{action execution}, asking whether a required action
was performed; \emph{landmark/target identification}, asking whether
an observed entity matches the intended landmark or goal; and
\emph{spatial/route relation}, asking whether a required spatial or
route relation held.

Across 3,862 queries, action execution appears in 68.44\%,
spatial/route relation in 55.72\%, and landmark/target identification
in 29.05\% (Table~\ref{tab:retrieval_query_dimensions}).
Furthermore, 48.24\% of queries involve at least two dimensions.
The generated questions thus frequently concern prior actions and
route relations, often jointly with other aspects of instruction
grounding.
This pattern is consistent with using episodic memory to assess
instruction execution, rather than only to recognize a destination.
The annotations characterize what the task executive asks to check.

\subsection{Qualitative Results and Real-World Demonstrations}
\label{sec:qualitative}

Figure~\ref{fig:result_vis} shows retrieval-supported recovery
in VLN and ObjectNav.
In the VLN example, retrieved transition observations leave
dining-room entry unconfirmed, prompting route reassessment.
The agent subsequently reaches the dining room and stops near
the white chair.
In the ObjectNav example, retrieved landmark observations
invalidate an accepted TV detection, reopening the search.
The agent then finds a real TV and terminates successfully.

\noindent\textbf{Real-World Navigation.}
To examine NavProbe's practicality beyond simulation, we deploy it on a Galaxea R1Pro equipped with a ZED2 RGB-D camera and a MID-360 LiDAR. Voxel-SLAM~\cite{voxelslam} provides pose estimation, Nav2~\cite{macenski2020marathon2} executes the metric goals selected by NavProbe. Subsequent observations inform task-state assessment and replanning. As shown in Figure~\ref{fig:real_world}, NavProbe successfully executes multi-stage real-world navigation instructions, including identifying visual landmarks and performing the corresponding navigation actions.

\section{Conclusion}

We presented NavProbe, a hierarchical zero-shot VLN agent that grounds
task-progress reasoning in selectively retrieved historical evidence.
Compact entity knowledge provides an index over a graph-structured
multimodal episodic memory, while the original visual and geometric
records remain available for inspection.
The agent can use these records to reassess subtask conditions and
completion states before selecting a navigation skill, and consolidate
reusable conclusions into memory for subsequent decisions.
Experiments on the evaluated R2R-CE and RxR-CE subsets demonstrate
strong navigation performance, while ablations support the contributions
of task-progress memory, evidence retrieval, and knowledge consolidation.
Retrieval analysis and a qualitative execution example illustrate how
historical observations support questions about instruction execution.
ObjectNav results and real-robot trials provide additional evidence of
the framework's applicability beyond the main VLN-CE evaluation.
Overall, the results support coupling compact navigation memory with
on-demand evidence access for explicit task-progress reasoning.

\bibliographystyle{IEEEtran}
\bibliography{references}

\end{document}